%% file: main.tex
\documentclass[fleqn,10pt]{wlscirep}
\usepackage[utf8]{inputenc}
\usepackage[T1]{fontenc}
\usepackage[numbers, sort]{natbib}
\usepackage{makecell}
\usepackage{multirow}
\title{Federated Learning for Heterogeneous Electronic Health Record Systems with Cost Effective Participant Selection}

\author[1]{Jiyoun Kim}
\author[1]{Junu Kim}
\author[1]{Kyunghoon Hur}
\author[1,*]{Edward Choi}
\affil[1]{KAIST, Kim Jaechul Graduate School of AI, Daejeon, 34141, Republic of Korea}

\affil[*]{edwardchoi@kaist.ac.kr}


\begin{abstract}
The increasing volume of electronic health records (EHRs) presents the opportunity to improve the accuracy and robustness of models in clinical prediction tasks. 
Unlike traditional centralized approaches, federated learning enables training on data from multiple institutions while preserving patient privacy and complying with regulatory constraints.
In practice, healthcare institutions (i.e., hosts) often need to build predictive models tailored to their specific needs (e.g., creatinine-level prediction, N-day readmission prediction) using federated learning.
When building a federated learning model for a single healthcare institution, two key challenges arise: (1) ensuring compatibility across heterogeneous EHR systems, and (2) managing federated learning costs within budget constraints.
Specifically, heterogeneity in EHR systems across institutions hinders compatible modeling, while the computational costs of federated learning can exceed practical budget limits for healthcare institutions.
To address these challenges, we propose EHRFL, a federated learning framework designed for building a cost-effective, host-specific predictive model using patient EHR data.
EHRFL consists of two components: (1) text-based EHR modeling, which facilitates cross-institution compatibility without costly data standardization, and (2) a participant selection strategy based on averaged patient embedding similarity to reduce the number of participants without degrading performance.
Our participant selection strategy sharing averaged patient embeddings is differentially private, ensuring patient privacy.
Experiments on multiple open-source EHR datasets demonstrate the effectiveness of both components.
With our framework, healthcare institutions can build institution-specific predictive models under budgetary constraints with reduced costs and time.
\end{abstract}
\begin{document}

\flushbottom
\maketitle
%
%
\thispagestyle{empty}

\section{Introduction}
With increasing daily interactions between patients and healthcare professionals, vast amounts of electronic health records (EHRs) are being accumulated.
Leveraging large-scale EHRs from multiple healthcare institutions can improve the accuracy and robustness of predictive models by allowing them to learn from diverse patient populations \citep{dash2019big, hur2023genhpf}.

Traditional approaches for multi-source EHR training have relied on centralized learning, where EHR data from different institutions are collected into a central repository.
However, this approach is often infeasible in clinical practice due to patient privacy concerns and regulatory constraints that restrict the transfer of patient data \citep{rieke2020future,rahman2023federated, teo2024federated}.
As an alternative, federated learning \citep{mcmahan2017communication} has emerged as a promising paradigm, enabling collaborative model construction across multiple local sites (i.e., clients) without sharing raw data.

In clinical practice, healthcare institutions often need to build a predictive model for their own specific targeted purposes (e.g., creatinine level prediction, N-day readmission prediction) using patient EHRs \citep{ashfaq2019readmission, davis2022effective, ghosh2021estimation}. 
In this case, healthcare institutions may wish to leverage data from other institutions to train their model.
Federated learning enables this by allowing a coordinating institution (i.e., the host) to collaborate with other institutions (i.e., subjects) that agree to participate—typically in exchange for compensation (e.g., data usage fees)—without directly accessing their raw data.

In such a scenario, the host faces two key challenges: (1) ensuring compatibility of EHR modeling across institutions with differing EHR systems \cite{hur2023genhpf}, and (2) managing overall federated learning costs to stay within budgetary constraints \cite{nguyen2021budget, white2023evaluating}.
Addressing both EHR heterogeneity and modeling cost constraints is critical for practical deployment of federated learning in healthcare \cite{hur2023genhpf, nguyen2021budget, white2023evaluating}.
To address these challenges, we propose EHRFL, a framework designed for cost-effective federated learning to build targeted predictive models on EHR data for the host institution. 
Our framework comprises the following two components.

First, our framework consists of text-based EHR federated learning, which enables compatibility across institutions with distinct EHR systems while maintaining cost-effectiveness.
Healthcare institutions often use different database schemas and medical coding systems (e.g., ICD-9, ICD-10) for storing patient records \citep{voss2015feasibility, glynn2019heterogeneity}.
To facilitate federated learning between the host and subjects, it is necessary to process the EHRs in a unified format.
One possible solution is to convert the host’s and subjects’ EHR systems to a common data model (CDM) format. 
However, this conversion is costly, time-consuming, and requires extensive involvement of multiple medical experts. 
Specifically, the conversion often requires thorough mapping of local EHR tables and clinical concepts (e.g., conditions, drugs, procedures) to predefined CDM tables and standard vocabularies \citep{biedermann2021standardizing, henke2024conceptual}.
In our framework, we adapt the recent work of \cite{hur2023genhpf}, which proposes a text-based conversion method for modeling heterogeneous EHR systems.
While their method demonstrates effectiveness in a centralized learning setting, we extend it to the federated learning setting.
By converting each institution’s sequence of EHR events into a corresponding textual format, this process enables compatible text-based federated learning across institutions of heterogeneous EHR systems, without extensive preprocessing or medical expert-driven, domain-specific mappings.

Second, we propose a method for selecting suitable federated learning participants based on averaged patient embedding similarity, to reduce the number of participants without compromising host model performance.
As the number of federated learning participants increases, the host incurs higher costs (e.g., data usage fees, computational costs) \citep{majeed2022factors}.
Thus, it is essential to effectively reduce the number of participating subjects without sacrificing the host's model performance.
Importantly, among the candidate subjects, certain subjects may not contribute positively or may even degrade the model performance for the host.
Identifying and excluding such subjects is crucial for cost-effective federated learning. 
To address this, our framework introduces a method for selecting participating subjects based on similarity between the host's and each subject's averaged patient embeddings.
To ensure patient privacy during the sharing averaged patient embeddings, we incorporate differential privacy \cite{dwork2006differential} when constructing averaged patient embeddings used for selecting participants.
Our approach aims to reduce the number of participating subjects while preserving host model performance, ultimately lowering the overall federated learning costs for the host.

In our experiments, we evaluate the two components of our framework.
First, we show that text-based EHR federated learning enables collaborative training between institutions with heterogeneous EHR systems, achieving performance gains while maintaining cost-effectiveness.
Second, we observe a correlation between performance gains for the host and average patient embedding similarity, supporting its use as a criterion for excluding subjects unlikely to contribute positively to the host model.
Third, we validate that averaged patient embedding similarity-based selection effectively reduces the number of participants without negatively impacting host model performance.
Our experimental results show that our framework enables the construction of an institution-specific model within budget constraints, with minimal cost and data-processing time.
As discussed in the Discussions section, our work is limited by the use of ICU data only, due to resource constraints. 
In future work, we plan to extend our framework to other clinical settings (e.g., outpatient data, emergency department data).
Overall, we believe our proposed EHRFL framework significantly enhances the practical applicability of federated learning in clinical settings, enabling healthcare institutions to build cost-effective, institution-specific predictive models within budget constraints.

\input{Tables/table4}
\input{Tables/table5}

\section{Overall Framework}

\begin{figure}[t] {\includegraphics[width=0.95\linewidth, clip, trim= 100 45 140 70]{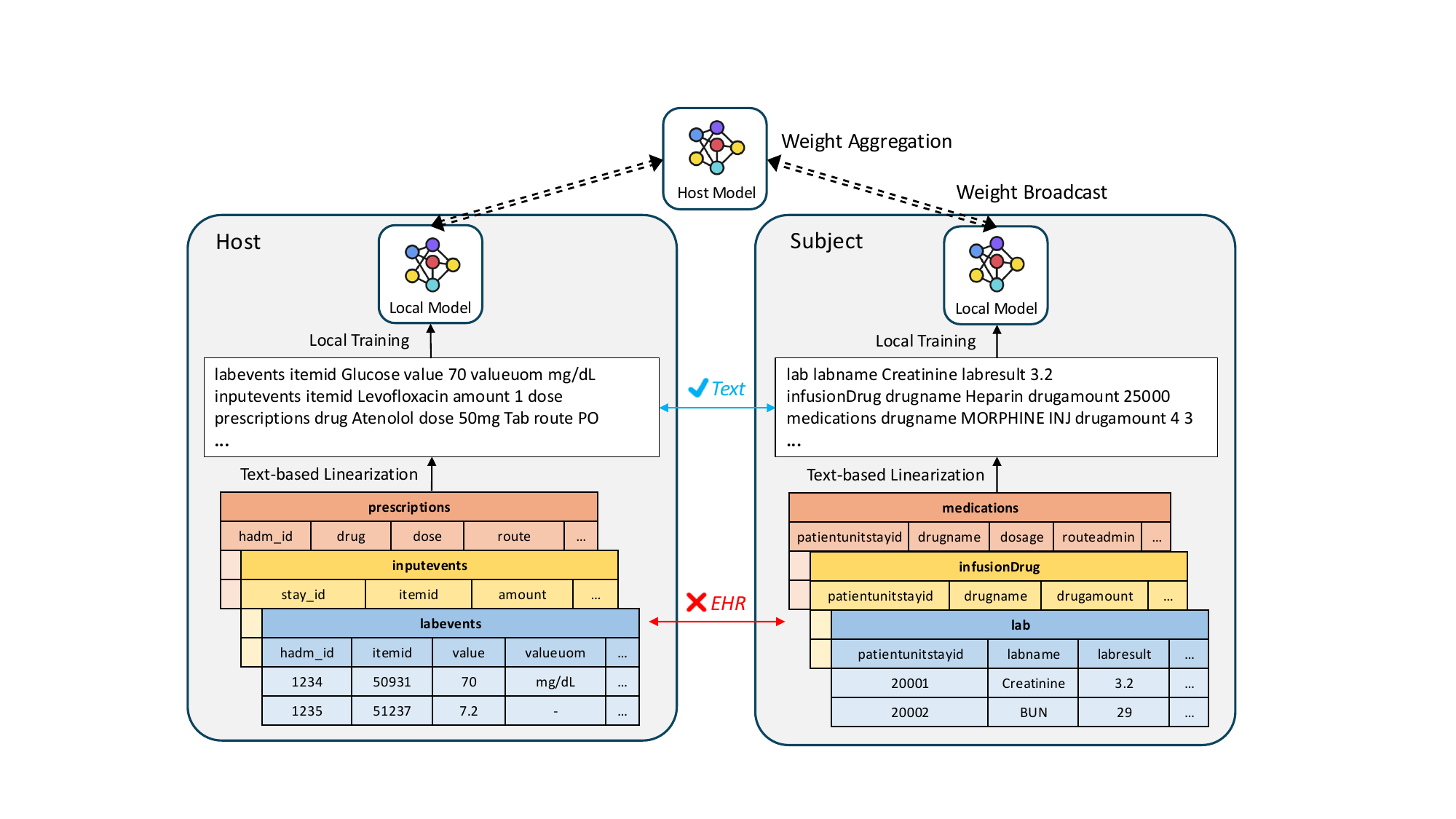}\vspace{-0.5em}}
  {\caption{Federated learning across healthcare institutions (i.e., host, subject) of heterogeneous EHR systems. EHR data is linearized into a standardized text-based format for compatible modeling.}\label{figure1}}
\end{figure}

\begin{figure}[t]
{\includegraphics[width=1\linewidth, clip, trim=10 55 10 140]{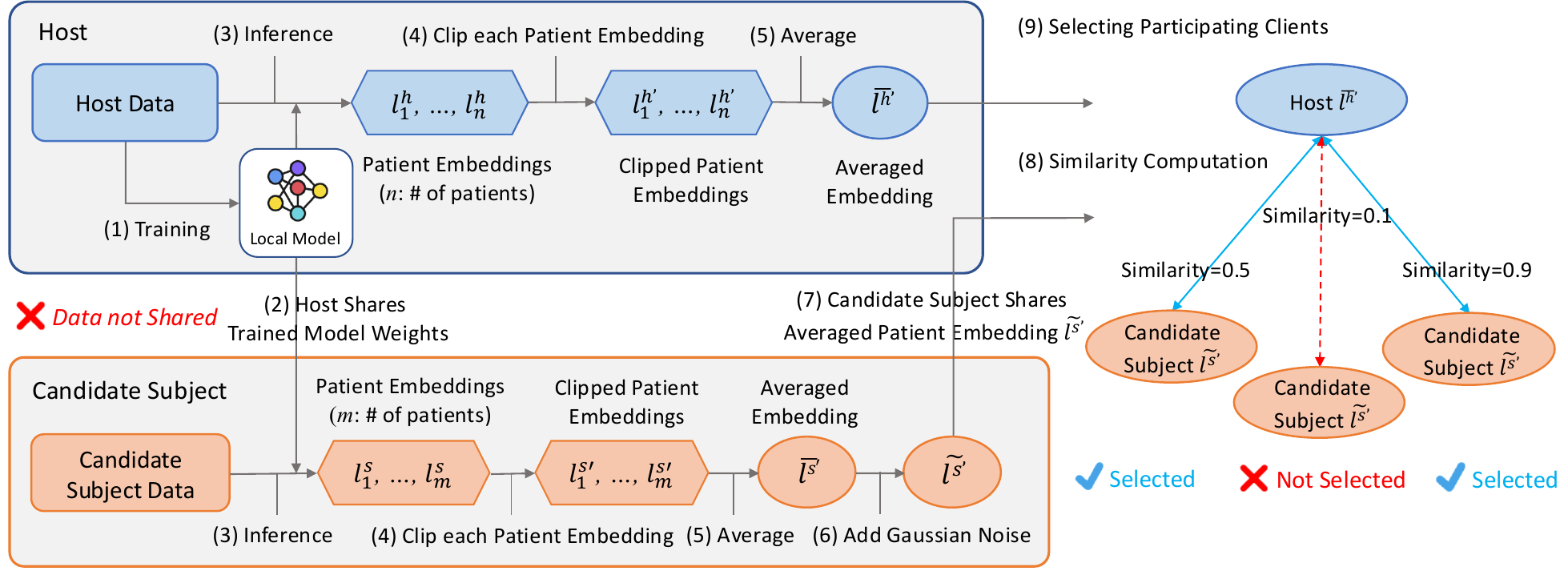}}
  {\caption{Selection of participating subjects in federated learning based on averaged patient embedding similarity with the host. To ensure privacy, each subject constructs its averaged patient embedding using differential privacy by clipping individual patient embeddings, averaging them, and adding Gaussian noise to the averaged embedding. To ensure consistency in similarity computation, the host applies the same clipping operation to its patient embeddings prior to averaging. Subjects with low similarity scores relative to the host are excluded from the federated learning process.}\label{figure2}}
\end{figure}

\begin{figure}[t] {\includegraphics[width=1\linewidth, clip, trim= 10 20 20 10]{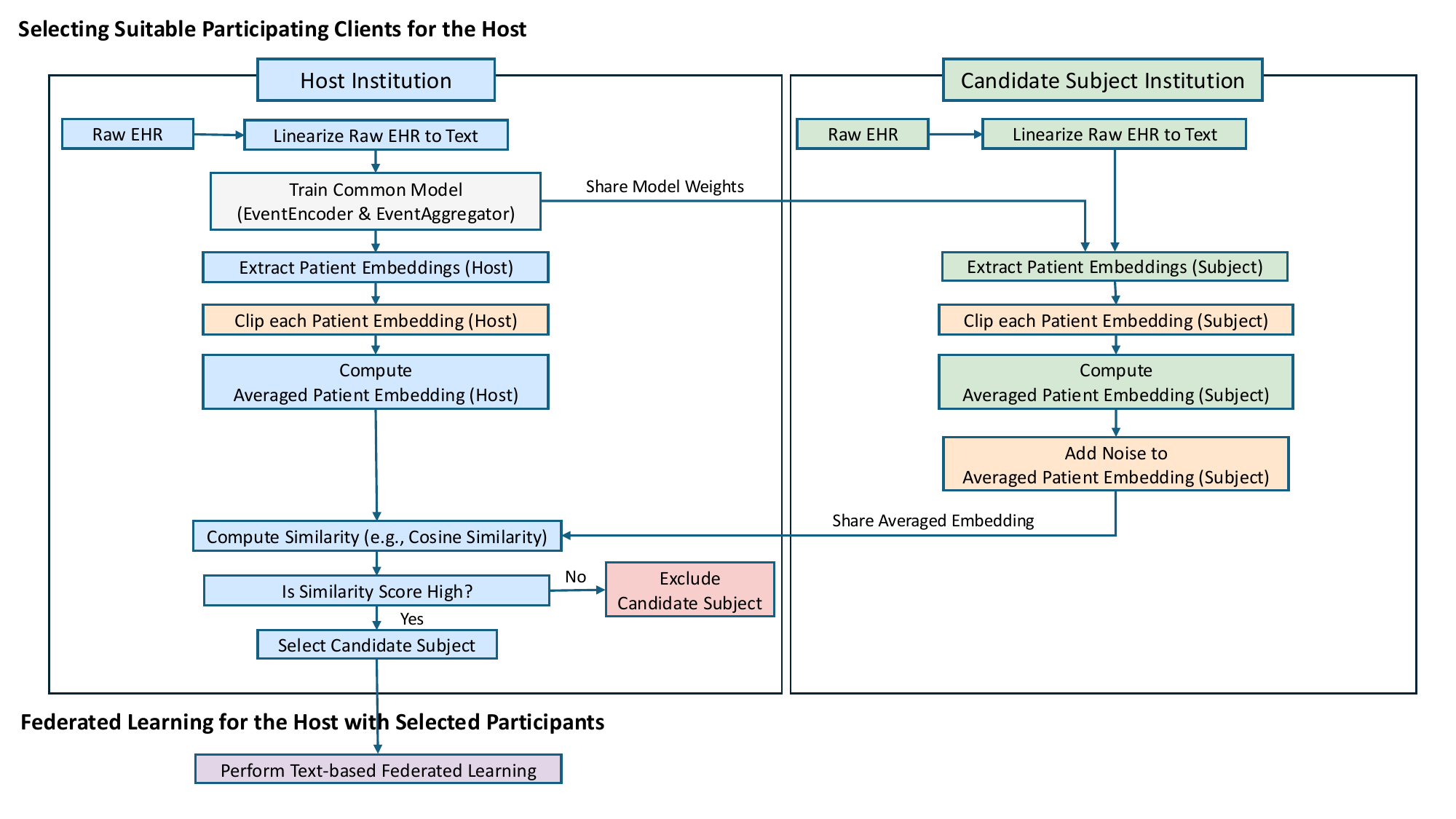}\vspace{-0.5em}}
  {\caption{UML Diagram for EHRFL}\label{figure3}}
\end{figure}

We introduce the two key components of our EHRFL framework.
The two components are (1) text-based EHR federated learning (Figure \ref{figure1}) and (2) participating client selection using averaged patient embeddings (Figure \ref{figure2}).
The overall UML diagram is shown in Figure \ref{figure3}.


\subsection{Text-based EHR Federated Learning}\label{text_based}
\subsubsection{Structure of EHRs}
The EHR data of a patient \textit{P} can be represented as a sequence of medical events $m_1$, $m_2$, ..., $m_k$, where \textit{k} denotes the total number of medical events for patient \textit{P}.
Each medical event $m_i$ (with \textit{i} as the event index), consists of an event type $e_i$ (e.g., `labevents', `inputevents', `prescriptions'), and a set of corresponding feature pairs \{($n_{i1}$, $v_{i1}$), ($n_{i2}$, $v_{i2}$), ...\}.
Each feature pair ($n_{ij}$, $v_{ij}$) includes a feature name $n_{ij}$ (e.g., `itemid', `value') and a feature value $v_{ij}$ (e.g., 50931 for `itemid', 70 for `value'). 
Given that different healthcare institutions may use distinct EHR schemas and medical code systems, variations may occur in event types, feature names, and feature values.
For instance, different institutions may use varying terminologies for event types or feature names, and distinct medical codes may be used to represent the same medical concept.
To enable federated learning across heterogeneous EHR systems, a compatible modeling format is required.

\subsubsection{Text-based EHR Linearization}
To achieve a standardized representation of EHRs across different healthcare institutions, each event $m_i$ of patient \textit{P} is converted into a text based format.
Specifically, we concatenate the event type $e_i$ with its corresponding feature pairs \{($n_{i1}$, $v_{i1}$), ($n_{i2}$, $v_{i2}$), ...\}, following the approach of \cite{hur2023genhpf}.
If feature value $v_{ij}$ is a medical code (e.g., 50931), it is replaced with its corresponding medical text (e.g., Glucose).
These code-to-text mappings (i.e., converting medical code $v_{ij}$ to its corresponding medical text) are typically stored as dictionaries in healthcare institution EHR databases, enabling automated conversion without requiring labor-intensive preprocessing or the involvement of medical experts.
The text-based medical event representation $r_i$ of medical event $m_i$ is represented as follows:

\begin{equation}
 r_i = ( e_i \oplus n_{i1} \oplus v_{i1}' \oplus n_{i2} \oplus v_{i2}' \cdots)
\end{equation}

where $\oplus$ denotes concatenation, and $v_{ij}'$ denotes the final feature value $v_{ij}$, either converted into its corresponding medical text or left unchanged if not a medical code.
Examples of these linearized events are shown in Figure \ref{figure1}, such as ``labevents itemid Glucose value 70 valueuom mg/dL''.

\subsubsection{Modeling of EHRs}\label{ehr_modeling}
For modeling the EHR of patient \textit{P}, we use a two-step encoding strategy based on \cite{hur2023genhpf}. 
First, each event representation $r_i$ is tokenized and encoded into an event vector $z_i$.
Let \textit{R} represent the event representations ($r_1$, $r_2$, ..., $r_k$) for patient \textit{P}, and \textit{Z} represent the corresponding event vectors ($z_1$, $z_2$, ..., $z_k$). 
This encoding process is perfomed using a two-layer Transformer-based event encoder model (i.e., \textit{EventEncoder}).

\begin{equation}
Z = EventEncoder(Tokenizer(R))
\end{equation}

Subsequently, the event vectors \textit{Z} of patient \textit{P} are further processed using a two-layer Transformer-based event aggregator model (i.e., \textit{EventAggregator}).
This model encodes the event vectors \textit{Z} and computes the average of the resulting vectors to produce the patient embedding \textit{l}.

\begin{equation}
l = EventAggregator(Z)
\end{equation}

For each target predictive task \textit{c} of the host, the patient embedding $l$ is put through a prediction head to generate the final task-specific prediction $\hat{l}_c$.
The model is trained by minimizing the sum of the prediction losses (e.g., cross-entropy loss) across all tasks $c_1$, ..., $c_t$, where \textit{t} denotes the total number of tasks that the host aims to predict.

\subsection{Participating Subject Selection using Averaged Patient Embeddings}\label{client_sel}

In this section, we present our method for selecting suitable participating subjects for the host using averaged patient embeddings. 
Motivated by prior research \cite{zhao2018federated, ozdayi2020improving, luo2023influence, li2020federated}, we hypothesize that subjects whose data distributions differ significantly from that of the host are more likely to have no positive impact or may even degrade the host model's performance.
To identify suitable subjects, we compute the similarity between the averaged patient embeddings extracted from the text-based EHR data of each candidate subject and that of the host.
To ensure patient privacy, we construct each subject’s averaged patient embedding using differential privacy.
Specifically, following prior differential privacy implementations, \cite{abadi2016deep,bu2023automatic,park2023differentially}, we clip individual patient embeddings, compute their average, and add Gaussian noise to the resulting averaged embedding.

\subsubsection{Subject Selection Process}\label{client_sel_steps}
The subject selection process proceeds as follows.
First, the host trains a model on its own data to build a common model for extracting patient embeddings.
This model is trained using the text-based EHR modeling method described in Section \ref{text_based}. 
This is to ensure that embeddings are generated within a shared embedding space, despite that host and subjects may have distinct EHR systems.
Second, the host shares the trained model weights to each candidate subject.
Third, the host and each candidate subject use the model to extract patient embeddings from their respective datasets.
Let $l^{h}_{i} \in \mathbb{R}^{d}$ denote the embedding of the $i$-th patient at the host, and $l^{s}_{j} \in \mathbb{R}^{d}$ denote the embedding of the $j$-th patient at a candidate subject.
The resulting embeddings for the host are denoted as
$\{l^{h}_{1}, l^{h}_{2}, \dots, l^{h}_{n}\}$,
while those for a candidate subject are denoted as
$\{l^{s}_{1}, l^{s}_{2}, \dots, l^{s}_{m}\}$,
where $n$ and $m$ represent the number of patients at the host and subject, respectively (note that $l$ is consistent with the notation in Section~\ref{ehr_modeling}).

Fourth, both the host and each candidate subject apply $\ell_{2}$-norm clipping to individual patient embeddings to bound the contribution of any single patient.
Specifically, each embedding is clipped to a maximum norm $C$ as follows:
\begin{equation}
l' = l \cdot \min\left(1, \frac{C}{\|l\|_2}\right).
\end{equation}
The resulting clipped patient embeddings for the host are denoted as ($l^{h'}_{1}$, $l^{h'}_{2}$, ..., $l^{h'}_{n}$), while the resulting clipped patient embeddings for a single candidate subject are denoted as ($l^{s'}_{1}$, $l^{s'}_{2}$, ..., $l^{s'}_{m}$).

Fifth, the host computes its averaged patient embedding from the clipped embeddings as
\begin{equation}
\bar{l}^{h'} = \frac{1}{n} \sum_{i=1}^{n} l^{h'}_{i},
\end{equation}
and each candidate subject similarly computes an averaged embedding
\begin{equation}
\bar{l}^{s'} = \frac{1}{m} \sum_{j=1}^{m} l^{s'}_{j}.
\end{equation}
This averaging produces a single embedding vector of dimensionality $d$, where $d$ is the dimensionality of each patient embedding.

Sixth, to ensure patient-level privacy, each candidate subject applies differential privacy to its averaged embedding before sharing it with the host.
Following the Gaussian mechanism, random Gaussian noise calibrated to sensitivity $C/m$ is added to $\bar{l}^{s'}$, yielding a differentially private averaged embedding:
\begin{equation}
\tilde{l}^{s'} = \bar{l}^{s'} + \mathcal{N}\left(0, \sigma^{2} \mathbf{I}\right),
\end{equation}
where
\begin{equation}
\sigma = \frac{C}{m} \cdot \frac{\sqrt{2 \log(1.25/\delta)}}{\epsilon},
\end{equation}
and $(\epsilon, \delta)$ are the differential privacy parameters.
In contrast, the host does not add noise to $\bar{l}^{h'}$, as the host embedding is not shared externally.
However, note that the same clipping operation (in the fourth step) is applied to both host and subject embeddings to ensure consistency in similarity computation.

Seventh, each candidate subject shares their averaged patient embedding, $\tilde{l}^{s'}$, to the host.
Eighth, for each candidate subject, the host computes the similarity between its own averaged embedding $\bar{l}^{h'}$ and the subject’s differentially private averaged embedding.
Similarity can be measured using metrics such as cosine similarity, euclidean distance, and KL divergence.
Ninth, the host excludes candidate subjects with low similarity scores, proceeding federated learning with subjects of high similarity scores that are not excluded.

A potential concern with this approach is that averaging patient embeddings across clients may result in information loss, which may lead to suboptimal selection of suitable clients.
Specifically, given that each client may have many patients, measuring similarity using averaged patient embeddings may lack the necessary granularity.
To address this concern, we additionally evaluate subject selection based on patient-wise similarities using non-averaged patient-level embeddings (Supplementary Tables S9, S10).
As detailed in Supplementary Tables S7-S10, our experiment results show that the averaged patient embedding method consistently leads to better host performance compared to selecting subjects using non-averaged embedding similarities.
Moreover, sharing patient-level embeddings across institutions raises privacy concerns, making the average embedding-based approach a reasonable method from both privacy and performance perspectives.

\subsection{Overall Cost Savings for the Host}

In this section, we analyze the cost savings achieved by using our framework.
Specifically, we compare the total cost incurred when all candidate subjects participate in federated learning versus the cost when only selected subjects participate, as determined by our selection framework.
The notations and total net savings are summarized in Table~\ref{tab:fl_costs}.

The total cost of using our framework includes:
(1) the cost of selecting suitable subjects using our averaged patient embedding similarity method, and
(2) the cost of performing federated learning with the selected subjects.
The cost of conducting federated learning with selected subjects ((2) above) consists of:
(i) the cost of converting EHR data into a text-based format,
(ii) data usage fees paid to each selected client,
(iii) computational costs for training.

As discussed in Section \ref{text_based}, unlike costly CDM-based conversion methods, text-based EHR modeling requires minimal cost, since it does not involve extensive preprocessing or expert-driven mappings. Conversion is done automatically using EHR database schema and code-to-text dictionaries within the EHR database.

Cost savings arise from avoiding data usage and computational costs for unselected clients. Specifically, as shown in Table~\ref{tab:fl_costs}, when only $K$ participants are involved instead of all $N$, both data usage fees and training costs are reduced proportionally by $(N - K)$.
Among the participating client selection costs, the dominant factors are the data usage fee $X$ and the training cost $C_{\text{train}}$, as the other components—$C_{\text{model}}$, $C_{\text{extract}}$, $C_{\text{average}}$, $C_{\text{embedding}}$, and $C_{\text{sim}}$—are associated with lightweight model inference or computation and are typically negligible in practice.

As shown in the total net savings equation, when $K \ll N$, cost savings are substantial since $X$ and $C_{\text{train}}$ scale with $(N - K)$.
Even in the minimal case where $K = N - 1$, the host still saves approximately $X + C_{\text{train}}$, indicating that our method provides cost benefits even under conservative selection scenarios.

\input{Tables/table0}

\section{Experimental Settings}

\subsection{Datasets}

For our experiments, we use three open-source EHR datasets: MIMIC-III \citep{johnson2016mimic}, MIMIC-IV \citep{johnson2023mimic}, and eICU \citep{pollard2018eicu}, each of which contains data from patients admitted to intensive care units (ICUs). 
MIMIC-III and MIMIC-IV are sourced from the same medical institution (i.e., Beth Israel Deaconess Medical Center) but differ in their database schemas and medical code systems (e.g., the itemid for `Dialysate Fluid' is 225977 in MIMIC-III, while it is 230085 in MIMIC-IV) and cover distinct time periods (MIMIC-III spans 2001 - 2012, MIMIC-IV spans 2008 - 2019).
Additionally, MIMIC-III includes data from two distinct clinical database systems (i.e., MetaVision, CareVue).
To reflect these differences, we set MIMIC-III-MV (MetaVision), MIMIC-III-CV (CareVue), and MIMIC-IV as three separate clients.
For MIMIC-IV, we exclude data from the 2008 - 2012 period to avoid overlap with MIMIC-III.
The eICU dataset contains data from hospitals across various regions of the United States.
To increase client diversity, we divide it into two clients, eICU-West and eICU-South, based on geographic location.

In summary, we construct a total of five clients, representing four distinct EHR database systems, covering three different geographical regions. 
Detailed demographic information and label distributions for each client are provided in Supplementary Tables S2 and S3.

\subsection{Setup}
\subsubsection{Cohort \& Prediction Tasks}
The cohort includes patients aged 18 years and older with ICU stays lasting at least 12 hours. 
For each hospital admission, only the first ICU stay is used.
Events occuring during the initial 12 hours of each ICU stay are used for prediction (i.e., 12 hour observation window).

We evaluate model performance on 12 clinical prediction tasks, following the tasks in \cite{hur2023genhpf}. 
A diverse set of tasks is included to demonstrate the effectiveness of our framework in building host models across various clinical scenarios.
The default prediction window is 24 hours, unless otherwise specified (see Supplementary Table S1 for detailed task descriptions).
Each model is trained using a multi-task learning, with 12 distinct prediction heads each corresponding to the individual tasks.
The data preprocessing code for cohort filtering and prediction task labeling is provided at: \url{https://github.com/Jwoo5/integrated-ehr-pipeline/tree/federated}.

\subsubsection{Differential Privacy Parameters}

Individual patient embeddings are clipped with an $\ell_2$-norm threshold of $C = 1.0$ prior to averaging.
The privacy budget is set to $\epsilon = 1.0$ and the failure probability to $\delta = 10^{-5}$.

\subsubsection{Federated Learning Algorithms}
Our proposed framework is algorithm-agnostic, making it compatible with different federated learning algorithms.
In the experiments, we use four widely used algorithms: FedAvg \citep{mcmahan2017communication}, FedProx \citep{li2020federated}, FedBN \citep{li2021fedbn}, FedPxN \citep{hwang2023towards}.

\subsubsection{Training Procedure}
The client datasets are each divided into training, validation, and test sets in an 8:1:1 ratio.
To ensure reproducibility and robustness of results, all experiments were conducted across three random seeds, including different data splits and model initializations.

For the federated learning experiments, we performed up to 300 communication rounds, with each client training for one local epoch per round.
For baseline comparison, we trained a model exclusively on the host's local data without federated learning (referred to as \textit{Single}) for up to 300 epochs.
In both settings, we applied an early stopping criterion: training would be halted if the macro AUROC, averaged across the 12 tasks, does not improve for 10 consecutive evaluations on the host's validation set.

Since the objective of our framework is to develop a model tailored specifically for the host, the final host model selected is the one that achieves the highest validation performance for the host across all communication rounds.
The experiments were conducted using four NVIDIA RTX 3090 GPUs with CUDA version 11.4, with each host model requiring approximately 1–2 days for training.
Detailed reproduction steps along with the full training code are provided in the following repository: \url{https://github.com/ji-youn-kim/EHRFL}.

\section{Results}

\input{Tables/table1}

\subsection{Federated Learning with Text-based EHR Modeling}
In this section, we evaluate the effectiveness of federated learning using text-based EHR modeling.
Table \ref{table1} shows a comparison of the host model performance when trained solely on its own data (i.e., \textit{Single}) versus when trained using federated learning with text-based EHR modeling. 
The standard deviation results for Table \ref{table1} are listed in Supplementary Table S4.
The performance of each host show that federated learning with text-based EHR modeling generally leads to improved host performance compared to training solely on local data (see bolded values).
These results indicate that text-based EHR federated learning enables leveraging data from multiple clients to improve the host's model performance, despite heterogeneity in EHR systems.
This trend is consistent across all of the four federated learning algorithms evaluated.
Importantly, text-based EHR modeling does not require extensive preprocessing or expert-driven domain-specific mappings, enabling low-cost federated learning when building a model for the host using heterogeneous EHR systems.

Another important observation is that increasing the number of participating clients does not always lead to better performance.
For example, when MIMIC-III-MV is the host, the \textit{Average} performance with three participating clients is 0.808, while the \textit{Best} performance with two participating clients is 0.812 .
This highlights the importance of selecting suitable clients for the host, indicating that simply increasing the number of participants may not always be beneficial.

\subsection{Relationship between Averaged Patient Embedding Similarity and Host Performance}\label{main:corr}

\input{Tables/table2}

In this section, we investigate the relationship between host-subject similarity, computed using averaged patient embeddings (as described in Section~\ref{client_sel}), and changes in host model performance when a subject participates in federated learning versus when it does not.
Identifying a correlation between the two would suggest that subjects with low similarity scores are more likely to not affect or degrade the host's performance, making it reasonable to exclude them from the federated learning process.

For measuring similarity, we use Cosine Similarity, Euclidean Distance, and KL Divergence. 
Cosine Similarity and Euclidean Distance are computed directly with the averaged patient embeddings $\bar{l}^h$ and $\bar{l}^s$.
For KL Divergence, we use the average across each softmax of the individual patient embeddings instead of the averaged patient embeddings.
In this case, during the embedding sharing phase in Section~\ref{client_sel_steps} (step five), each candidate subject shares the averaged softmax values instead of the averaged patient embeddings, consistently ensuring privacy by avoiding exchange of patient-level data.

Table \ref{table2} presents the observed correlations between (1) similarity between the host and a subject, measured using averaged patient embeddings (or averaged softmax values for KL Divergence) and (2) the difference in host performance with and without the subject participating in federated learning (if multiple subjects participate, other subjects consistently participate in both scenarios to isolate the effect of the target subject).
The results show consistent correlations across all three similarity methods and the different numbers of participating clients.
This trend holds across each of the four federated learning algorithms (see Supplementary Table S5).
Note that Cosine Similarity exhibits positive correlations (as higher values indicate greater similarity), while Euclidean Distance and KL Divergence exhibit negative correlations (as higher values indicate lower similarity). 
These correlations indicate that averaged patient embedding similarity can serve as a reliable criterion for selecting which subjects to exclude from federated learning, particularly by eliminating those with lower similarity scores (or larger distances).

\subsection{Selecting Participating Subjects using Averaged Patient Embedding Similarity}

\input{Tables/table3}

Based on the correlations observed in Section~\ref{main:corr}, we apply our proposed method of selecting participating subjects using averaged patient embedding similarity.
Table \ref{table3} presents the host performance for different numbers of participating clients. For each case, subjects with lower similarity scores (or larger distances) are excluded, retaining only a fixed number of participants.
The results show that in 13 out of 15 cases, selecting clients based on averaged patient embedding similarity achieves performance equal to or better than using all five clients, across all similarity metrics (see underlined values).
This demonstrates that our method enables reducing the number of participants without compromising host model performance.

Furthermore, in 8 out of 15 cases, our method achieves performance equal to the \textit{Best} case (see bolded values), and in 14 out of 15 cases, it exceeds the \textit{Average} performance (see asterisked values).
These comparisons with the \textit{Best} and \textit{Average} baselines further validate the effectiveness of averaged patient embedding similarity as a criterion for selecting suitable clients.

Detailed performance results for each individual federated learning algorithm are presented in Supplementary Tables S7 and S8.
These results show that our selection method consistently achieves performance equal to or better than using all 5 clients and consistently matches the \textit{Best} and exceeds the \textit{Average} performance in most cases across each of the four algorithms.
Overall, these results highlight the effectiveness of our averaged patient embedding-based method in identifying suitable subjects for the host, thereby enabling effective reduction in the number of participating clients without sacrificing host model performance.

\section*{Discussion}
In this work, we address two practical challenges that arise when building federated learning models tailored to a single healthcare institution: (1) heterogeneity across EHR systems and (2) budget constraints associated with building the federated learning model.
As highlighted in prior work \cite{hur2023genhpf}, differences in EHR schemas and medical coding systems pose a fundamental barrier to collaborative modeling across healthcare institutions. 
At the same time, recent studies emphasize that cost considerations are critical for real-world adoption of modeling in healthcare, particularly under constrained institutional budgets \cite{nguyen2021budget, white2023evaluating}. 
Our proposed EHRFL framework jointly addresses these challenges by enabling compatible text-based EHR modeling while introducing a cost-aware participant selection strategy in federated learning.

\subsection*{Assumptions}

As is common in healthcare institutions, our framework is designed for a cross-silo federated learning setting, where participating institutions are assumed to have relatively stable network connectivity.
Under this assumption, the communication of model parameters and averaged embeddings does not constitute a major bottleneck, in contrast to cross-device federated learning scenarios characterized by unstable or bandwidth-limited connections. 

\subsection*{Criteria for Selecting Suitable Participating Clients}

One of the core components of our framework is the use of averaged patient embedding similarity as a criterion for selecting suitable federated learning participants. Our experimental results show that using our averaged patient embedding similarity method enables identifying and excluding subjects that contribute marginally or negatively, effectively reducing federated learning costs without compromising performance. 
The use of embedding-based dataset representations aligns with prior research on dataset similarity and representation learning \cite{helali2021scalable, dias2021imagedataset2vec, van2021effectiveness}, while extending these ideas to a privacy-preserving, federated healthcare setting. 
Although embedding similarity is effective in our experiments, alternative criteria—such as label distributions, feature statistics, or performance-based metrics—could also inform participant selection \cite{stolte2024methods, elhussein2024universal, leite2021exploiting}. 
A systematic comparison of different selection criteria may further improve robustness, which we leave for future work as this is the first work for participating client selection. 

\subsection*{Scalability of Selecting Suitable Participating Clients}
Our participant selection strategy requires training the embedding model only once, which is performed by the host. The trained model weights are then shared with candidate subjects, who use the model solely for embedding extraction and do not perform any model training. As a result, the per-subject overhead is limited to lightweight inference for embedding extraction and the communication of a single averaged embedding. Even when the number of candidate institutions is large, the dominant computational cost—model training—remains constant, while communication and inference costs scale linearly and remain modest.

\subsection*{Limitations and Future Works}

A limitation of our work is that our experiments is restricted to ICU datasets, due to resource constraints. 
Although we incorporate multiple heterogeneous ICU datasets spanning different EHR systems, further validation on other clinical settings—such as outpatient, emergency department data—can be important to assess the generalizability of our approach. 
In addition, our framework does not explicitly incorporate data quality checks or data cleaning procedures during the data preprocessing stage. Integrating data quality control and data cleaning strategies, as studied in prior research \citep{mavrogiorgos2022multi,mavrogiorgou2019iot, leema2011effective}, may help remove erroneous data and further improve the robustness and reliability of the text representations. We leave such applications to future work.

\subsection*{Broader Applicability beyond Healthcare}

Beyond healthcare, the principles underlying EHRFL may generalize to other domains in which a single institution aims to build a predictive model for their needs with other institutions under schema heterogeneity and cost constraints. 
For example, in retail and e-commerce forecasting, organizations often operate heterogeneous point-of-sale and inventory systems \cite{kholod2024analysis}, and participant selection based on consumer behavioral similarity could reduce collaboration costs without exposing sensitive transaction data. 
Similarly, in manufacturing and supply-chain predictive maintenance, factories often use different Industrial IoT platforms and sensor systems.
Sensor platforms and maintenance logs could be linearized into text format regardless of the underlying data schema, and manufacturers could select federated learning participants with similar operating conditions. 
These extensions suggest that the proposed framework may have broader applicability beyond clinical prediction tasks.

\section*{Conclusion}

In this paper, we introduce EHRFL, a federated learning framework for developing a client-specific model using EHRs. By incorporating text-based EHR modeling, EHRFL enables federated learning across clients with heterogeneous EHR systems. 
Furthermore, our framework optimizes the selection of participating clients using averaged patient embeddings, reducing the number of participants without compromising model performance.

This work can benefit multiple stakeholders in the healthcare ecosystem. Healthcare institutions can leverage EHRFL to build predictive models tailored to their internal needs while controlling federated learning costs and preserving patient privacy. Clinical data departments and can adopt the text-based EHR federated learning approach to enable collaborative modeling across incompatible EHR systems with minimal preprocessing overhead. 
From a research perspective, EHRFL provides a practical framework for studying federated learning under realistic healthcare constraints, including heterogeneity, privacy, and budget limitations. 
We release open-source code and experimental details to support reproducibility and adoption.

As next steps, we plan to extend EHRFL in several directions. 
First, we aim to evaluate the framework on additional clinical settings beyond the ICU, such as outpatient and emergency department data, to further assess generalizability. Second, we plan to incorporate data quality control mechanisms prior to text-based linearization to improve robustness in real-world deployments. 
Third, we will explore alternative client selection criteria to compare and enhance the selection robustness. 
These extensions can be addressed incrementally in future work, with a focus on improving generalizability, robustness, and evaluation breadth. 
We believe these steps provide a sustainable path toward deploying cost-effective federated learning systems in practical healthcare environments.

\bibliography{sample}

\section*{Funding}
This work was supported by the Institute of Information \& Communications Technology Planning \& Evaluation (IITP) grant (No.RS-2019-II190075), and National Research Foundation of Korea (NRF) grant (NRF-2020H1D3A2A03100945), funded by the Korea government (MSIT).

\section*{Author contributions statement}

J.K.: Designed the method, conducted experiments, and wrote the manuscript. J.K.: provided discussions and feedback, and reviewed the manuscript. K.H.: provided discussions. E.C.: provided supervisions and analysis, and reviewed the manuscript. 

\section*{Data availability}
This paper uses MIMIC-III (\url{https://physionet.org/content/mimiciii/1.4/}), MIMIC-IV (\url{https://physionet.org/content/mimiciv/2.0/}), and eICU (\url{https://physionet.org/content/eicu-crd/2.0/}) datasets, which are available on Physionet with Physionet credentials.

\section*{Code availability}
The source code used in this research is publicly available on Github: \url{https://github.com/ji-youn-kim/EHRFL}.

\section*{Additional information}
The author(s) declare no competing interests.

\end{document}

%% file: Tables/table4.tex
\begin{table}[t]
\centering
\caption{Comparison with prior EHR federated learning studies. 
A common limitation of prior work is the oversimplification of federated learning client construction, where a single EHR dataset is artificially partitioned to simulate multiple clients, or multiple hospitals with identical EHR schemas are used, failing to reflect real-world EHR heterogeneity.}
\resizebox{\columnwidth}{!}{
\begin{tabular}{l l l c}
\toprule
\textbf{Work} & \textbf{Objective} & \textbf{Dataset(s)} & \textbf{Dataset Heterogeneity} \\
\midrule
\cite{lee2020federated} 
& In-hospital mortality prediction 
& MIMIC-III 
& No \\

\cite{vaid2021federated} 
& 7-day mortality prediction for COVID-19 patients 
& Mount Sinai Health System 
& No \\

\cite{dang2022federated} 
& In-hospital mortality and AKI prediction in ICU patients 
& eICU 
& No \\

\cite{rajendran2023data} 
& AKI and sepsis risk prediction in ICU 
& eICU 
& No \\

\textbf{Ours} 
& Institution-specific clinical prediction modeling 
& MIMIC-III (CV/MV), MIMIC-IV, eICU (South/West) 
& Yes \\
\bottomrule
\end{tabular}
}
\end{table}

%% file: Tables/table5.tex
\begin{table}[t]
\centering
\caption{Comparison with prior studies using client similarity measurement in federated learning. Unlike our work, prior studies focus on constructing models optimized for multiple clients simultaneously (e.g., clustered federated learning), rather than building a model targeted for a single institution.}
\small
\begin{tabular}{l l l}
\toprule
\textbf{Work} & \textbf{Similarity Metric} & \textbf{Target Clients} \\
\midrule
\cite{huang2019patient} & K-Means Clustering & Multiple Institutions \\
\cite{elhussein2023privacy} & Spectral Clustering & Multiple Institutions \\
Ours & Cosine Similarity, Euclidean Distance, KL-Divergence & Single Institution \\
\bottomrule
\end{tabular}
\label{tab:similarity_comparison}
\end{table}

%% file: Tables/table0.tex
\begin{table}[ht]
\centering
\caption{Notation and cost savings in using our participating subject selection method. The total net savings are computed by (Data usage fee savings) + (Computational cost savings) - (Participating client selection costs). Data usage fee savings and computational cost savings are computed by comparing costs when all N candidates participate, to when only a selected K number of subjects participate in federated learning.}
\label{tab:fl_costs}
\begin{tabular}{ll}
\toprule
\textbf{Symbol / Term} & \textbf{Definition} \\
\midrule
$N$ & Total number of candidate clients in FL (including the host) \\
$K$ & Number of selected participating clients in FL (including the host) \\
$X$ & Data usage fee for a single hospital \\
$R$ & Number of training communication rounds in FL \\
$L$ & Number of local epochs trained before each communication round in FL \\
$E$ & Number of epochs for training a local model \\
$C_{\text{train}}$ & Training cost for a client training the model for 1 local epoch \\
$C_{\text{model}}$ & Transmission cost for sending model weights between servers \\
$C_{\text{extract}}$ & Cost for extracting embeddings from the host's pre-trained model \\
$C_{\text{average}}$ & Cost for averaging individual embeddings into a single embedding \\
$C_{\text{embedding}}$ & Transmission cost for sending embeddings between servers \\
$C_{\text{sim}}$ & Cost for computing similarity between two embeddings \\
\midrule
\multicolumn{2}{l}{\textbf{Participating Client Selection Costs}} \\
Local model training costs & $E \cdot C_{\text{train}}$ \\
Transmitting model candidates & $N \cdot C_{\text{model}}$ \\
Extracting embeddings & $N \cdot C_{\text{extract}}$ \\
Averaging embeddings & $N \cdot C_{\text{average}}$ \\
Transmitting embeddings to the host & $N \cdot C_{\text{embedding}}$ \\
Computing similarities with the host & $N \cdot C_{\text{sim}}$ \\
Overall Costs & $E \cdot C_{\text{train}} + N \cdot (C_{\text{model}} + C_{\text{extract}} + C_{\text{average}} + C_{\text{embedding}} + C_{\text{sim}})$ \\
\midrule
\multicolumn{2}{l}{\textbf{Data Usage Fee Savings}} \\
Overall Savings & $(N - K) \cdot X$ \\
\midrule
\multicolumn{2}{l}{\textbf{Computational Cost Savings}} \\
Local training cost savings & $(N - K) \cdot R \cdot L \cdot C_{\text{train}}$ \\
Transmission (client to server) savings & $(N - K) \cdot R \cdot C_{\text{model}}$ \\
Transmission (server to client) savings & $(N - K) \cdot R \cdot C_{\text{model}}$ \\
Overall Savings & $(N - K) \cdot R \cdot (L \cdot C_{\text{train}} + 2 \cdot C_{\text{model}})$ \\
\midrule
\textbf{Total Net Savings} & 
\makecell[l]{
$(N - K) \cdot X + ((N - K) \cdot R \cdot L - E) \cdot C_{\text{train}}$\\
$+ (2 \cdot (N - K) \cdot R - N) \cdot C_{\text{model}}$\\
$- N \cdot (C_{\text{extract}} + C_{\text{average}} + C_{\text{embedding}} + C_{\text{sim}})$
}
\\
\bottomrule
\end{tabular}
\end{table}

%% file: Tables/table1.tex
\begin{table*}[ht]
\centering
\setlength{\tabcolsep}{2pt}
\caption{Comparison of the host performance (average macro AUROC of the 12 tasks) trained solely on local data (i.e., \textit{Single}) versus trained on federated learning using text-based EHR modeling with a fixed number of clients.
The standard deviation results are listed in Supplementary Table S4. 
Each client column represents the performance when the client is the host. \textit{Average} and \textit{Best} represent the average and best host performance across all subject combinations for the fixed number of clients. Bold indicates federated learning performance higher than \textit{Single} results.} \label{table1}
\begin{tabular}{c|c|cc|cc|cc|cc|cc}
\toprule
\textbf{\# Clients} & \textbf{Method}  & \multicolumn{2}{c|}{\textbf{MIMIC-III-MV}} &  \multicolumn{2}{c|}{\textbf{MIMIC-III-CV}} & \multicolumn{2}{c|}{\textbf{MIMIC-IV}} & \multicolumn{2}{c|}{\textbf{eICU-West}} & \multicolumn{2}{c}{\textbf{eICU-South}}  \\ \midrule
1          & Single & \multicolumn{2}{c|}{0.802} & \multicolumn{2}{c|}{0.799} & \multicolumn{2}{c|}{0.814} & \multicolumn{2}{c|}{0.77} & \multicolumn{2}{c}{0.775}  \\ \midrule 
- &  -       & Average        & Best       & Average        & Best       & Average        & Best       & Average        & Best       & Average        & Best       \\ \midrule
\multirow{5}{*}{2} & FedAvg    & \textbf{0.806}      & \textbf{0.812}   & 0.791 & 0.797   & 0.814      & \textbf{0.815}       & \textbf{0.779}      & \textbf{0.786}     & 0.774      & \textbf{0.778}      \\
           & FedProx    & \textbf{0.804}      & \textbf{0.812}   & 0.792     & \textbf{0.800}      & 0.812      & 0.814       & \textbf{0.779}      & \textbf{0.785}    & 0.772      & \textbf{0.777}       \\
           & FedBN     & \textbf{0.808}      & \textbf{0.811}   & 0.793      & 0.799       & 0.814      & \textbf{0.816}       & \textbf{0.779}      & \textbf{0.787}     & 0.775      & \textbf{0.779}      \\
           & FedPxN    & \textbf{0.807}      & \textbf{0.813}   & 0.794     & \textbf{0.800}       & 0.812      & \textbf{0.815}     & \textbf{0.779}      & \textbf{0.785}       & 0.774      & \textbf{0.777}      \\ 
           & Avg.        & \textbf{0.806}      & \textbf{0.812}   & 0.792      & 0.799      & 0.813      & \textbf{0.815}      & \textbf{0.779}      & \textbf{0.786}    & 0.774      & \textbf{0.778}        \\ \midrule
\multirow{5}{*}{3}          & FedAvg     & \textbf{0.808}      & \textbf{0.814}   & 0.793      & \textbf{0.800}      & 0.812      & \textbf{0.817}     & \textbf{0.780}      & \textbf{0.786}      & \textbf{0.776}      & \textbf{0.780}       \\
           & FedProx   & \textbf{0.808}      & \textbf{0.814}   & 0.793     & \textbf{0.800}       & 0.812      & \textbf{0.816}      & \textbf{0.780}      & \textbf{0.784}      & 0.775      & \textbf{0.780}      \\
           & FedBN      & \textbf{0.808}      & \textbf{0.815}    & 0.795     & \textbf{0.800}     & 0.813      & \textbf{0.817}      & \textbf{0.782}      & \textbf{0.787}     & \textbf{0.777}      & \textbf{0.781}       \\
           & FedPxN    & \textbf{0.808}      & \textbf{0.814}   & 0.795     & \textbf{0.800}       & 0.813      & \textbf{0.816}       & \textbf{0.781}      & \textbf{0.787}    & 0.775      & \textbf{0.781}       \\ 
           & Avg.       & \textbf{0.808}      & \textbf{0.814}   & 0.794      & \textbf{0.800}       & 0.812      & \textbf{0.816}      & \textbf{0.781}      & \textbf{0.786}     & \textbf{0.776}      & \textbf{0.781}       \\ \midrule
\multirow{5}{*}{4}          & FedAvg    & \textbf{0.810}      & \textbf{0.813}   & 0.793      & 0.797       & 0.813      & \textbf{0.817}        & \textbf{0.782}      & \textbf{0.786}     & \textbf{0.776}      & \textbf{0.780}     \\
           & FedProx   & \textbf{0.809}      & \textbf{0.813}    & 0.792      & 0.798      & 0.812      & \textbf{0.815}      & \textbf{0.782}      & \textbf{0.786}      & \textbf{0.776}      & \textbf{0.779}      \\
           & FedBN     & \textbf{0.812}      & \textbf{0.818}   & 0.796      & 0.799       & 0.814      & \textbf{0.816}      & \textbf{0.783}      & \textbf{0.788}     & \textbf{0.778}      & \textbf{0.780}       \\
           & FedPxN    & \textbf{0.812}      & \textbf{0.816}   & 0.796      & \textbf{0.801}       & 0.812      & \textbf{0.816}        & \textbf{0.784}      & \textbf{0.790}     & \textbf{0.777}      & \textbf{0.780}     \\
           & Avg.       & \textbf{0.811}      & \textbf{0.815}   & 0.794      & 0.799       & 0.813      & \textbf{0.816}       & \textbf{0.783}      & \textbf{0.788}     & \textbf{0.777}      & \textbf{0.780}      \\ \midrule
\multirow{5}{*}{5}          & FedAvg    & \textbf{0.812}      & \textbf{0.812}   & 0.795      & 0.795       & 0.812      & 0.812       & \textbf{0.782}      & \textbf{0.782}     & \textbf{0.780}      & \textbf{0.780}      \\
           & FedProx    & \textbf{0.810}      & \textbf{0.810}   & 0.791      & 0.791      & 0.811      & 0.811    & \textbf{0.780}      & \textbf{0.780}      & 0.773      & 0.773        \\
           & FedBN     & \textbf{0.815}      & \textbf{0.815}   & 0.798      & 0.798       & 0.815      & \textbf{0.815}       & \textbf{0.785}      & \textbf{0.785}      & \textbf{0.780}      & \textbf{0.780}     \\
           & FedPxN    & \textbf{0.813}      & \textbf{0.813}   & 0.798      & 0.798       & 0.813      & 0.813      & \textbf{0.783}      & \textbf{0.783}     & \textbf{0.777}      & \textbf{0.777}       \\
           & Avg.       & \textbf{0.813}      & \textbf{0.813}   & 0.796      & 0.796       & 0.813      & 0.813      & \textbf{0.782}      & \textbf{0.782}    & \textbf{0.777}      & \textbf{0.777}        \\ \bottomrule
\end{tabular}
\end{table*}

%% file: Tables/table2.tex
\begin{table*}[ht]
\centering
\caption{Correlation between host-subject similarity and the change in host performance when a subject participates in federated learning compared to when the subject does not. The correlations are averaged across that of FedAvg, FedProx, FedBN, and FedPxN. Similarity is measured using averaged patient embeddings (or averaged softmax for KL divergence). See Supplementary Table S5 for details.
}\label{table2}

\begin{tabular}{llccccc}
\toprule
\textbf{Method}   & \textbf{Metric} & \textbf{2 Clients} & \textbf{3 Clients} & \textbf{4 Clients} & \textbf{5 Clients} & \textbf{Overall} \\
\midrule

Cosine    & Spearman & 0.453  & 0.452  & 0.415  & 0.394  & 0.425  \\
          & Kendall  & 0.323  & 0.314  & 0.288  & 0.282  & 0.296  \\
Euclidean & Spearman & -0.489 & -0.453 & -0.404 & -0.362 & -0.425 \\
          & Kendall  & -0.343 & -0.314 & -0.278 & -0.258 & -0.292 \\
KL        & Spearman & -0.514 & -0.444 & -0.390 & -0.329 & -0.415 \\
          & Kendall  & -0.364 & -0.311 & -0.268 & -0.230 & -0.288 \\
\bottomrule
\end{tabular}
\end{table*}

%% file: Tables/table3.tex
\begin{table*}[ht]
\centering
\setlength{\tabcolsep}{3pt}
\caption{Host performance (average macro AUROC across 12 tasks) when selecting participating subjects using averaged patient embeddings (Cosine Sim., Euclidean Dist., and KL Div.), averaged across FedAvg, FedProx, FedBN, FedPxN.
The performance is compared to (1) using all 5 clients, (2) \textit{Average}, and (3) \textit{Best}. \textit{Average} and \textit{Best} refer to the average and best host performance, respectively, across all subject combinations for a fixed number of participating clients. Underlined values indicate performance equal to or better than when using all 5 clients, bold indicates performance equal to \textit{Best} performance, asterisk indicates performance exceeding \textit{Average} performance. Standard deviations are listed in Supplementary Table S6.}\label{table3}
\begin{tabular}{c|c|c|c|c|c|c}
\toprule
\textbf{\# Clients} & \textbf{Metric}     & \textbf{MIMIC-III-MV} & \textbf{MIMIC-III-CV} & \textbf{MIMIC-IV} & \textbf{eICU-West} & \textbf{eICU-South}  \\ \midrule
\multirow{5}{*}{2}          & Average      & 0.806   & 0.792           & 0.813     & 0.779    & 0.774       \\ \cmidrule{2-7}
           & Cosine           & 0.810*    & \textbf{\underline{0.799}}*      & \underline{0.814}*    & \textbf{\underline{0.784}}*   & \textbf{\underline{0.778}}*         \\ 
           & Euclidean      & 0.810*   & \textbf{\underline{0.799}}*         & \underline{0.814}*     & \textbf{\underline{0.784}}*  & \textbf{\underline{0.778}}*         \\ 
           & KL            & 0.810*     & \textbf{\underline{0.799}}*        & \underline{0.814}*    & \textbf{\underline{0.784}}*    & \textbf{\underline{0.778}}*        \\ \cmidrule{2-7}
           & Best          & 0.811    & 0.799         & 0.815    & 0.784    & 0.778        \\ \midrule
\multirow{5}{*}{3}          & Average      & 0.808     & 0.794         & 0.812       & 0.781   & 0.776     \\ \cmidrule{2-7}
           & Cosine        & \textbf{\underline{0.813}}*   & \textbf{\underline{0.799}}*          & \underline{0.814}*   & \underline{0.783}*     & \underline{0.777}*        \\ 
           & Euclidean      & \textbf{\underline{0.813}}*  & \textbf{\underline{0.799}}*          & \underline{0.814}*   & \underline{0.783}*     & \underline{0.778}*        \\ 
           & KL              & \textbf{\underline{0.813}}*   & \textbf{\underline{0.799}}*        & \underline{0.814}*   & \underline{0.783}*     & \underline{0.777}*        \\ \cmidrule{2-7}
           & Best          & 0.813   & 0.799          & 0.816    & 0.786    & 0.779        \\ \midrule
\multirow{5}{*}{4}          & Average       & 0.811    & 0.794         & 0.813     & 0.783   & 0.777        \\ \cmidrule{2-7}
           & Cosine        & \textbf{\underline{0.814}}*    & \underline{0.796}*         & 0.812   & \textbf{\underline{0.787}}*    & \textbf{\underline{0.778}}*         \\ 
           & Euclidean       & \textbf{\underline{0.814}}*    & \underline{0.796}*       & 0.812   & \textbf{\underline{0.787}}*   & \textbf{\underline{0.778}}*          \\ 
           & KL             & \textbf{\underline{0.814}}*     & \underline{0.796}*       & 0.812    & \textbf{\underline{0.787}}* & \textbf{\underline{0.778}}*           \\ \cmidrule{2-7}
           & Best           & 0.814     & 0.797       & 0.816  & 0.787   & 0.778           \\ \midrule
5          & -            & 0.813    & 0.796          & 0.813    & 0.782  & 0.777          \\ \bottomrule
\end{tabular}
\label{tab:metrics}
\end{table*}